\documentclass[letterpaper, 10 pt, conference]{ieeeconf}  % Comment this line out if you need a4paper

\IEEEoverridecommandlockouts                              % This command is only needed if 
\usepackage{graphics} % for pdf, bitmapped graphics files
\usepackage{epsfig} % for postscript graphics files
\usepackage{times} % assumes new font selection scheme installed
\usepackage{amsmath} % assumes amsmath package installed
\usepackage{amssymb}  % assumes amsmath package installed
\usepackage{color}
\usepackage{algorithm}%
\usepackage{algorithmicx}%
\usepackage{algpseudocode}%
\usepackage{threeparttable}
\usepackage{multirow}
\usepackage{graphicx}
\usepackage{subfigure}

\usepackage{cite}
\usepackage[dvipsnames]{xcolor}
\usepackage{adjustbox}
\usepackage{array}
\usepackage{makecell}
\usepackage{url}

\definecolor{paleorange}{rgb}{0.85, 0.45, 0.1} % darker orange-brown
\definecolor{skyblue}{rgb}{0.1, 0.45, 0.8}     % darker blue
\definecolor{palegreen}{rgb}{0.1, 0.6, 0.3}    % darker green

\title{\LARGE \bf
}

\title{xFODE: An Explainable Fuzzy Additive ODE Framework for System Identification \\
\thanks{This work was supported by MathWorks\textsuperscript{\textregistered} in part by a Research Grant awarded to T. Kumbasar. Any opinions, findings, conclusions, or recommendations expressed in this paper are those of the authors and do not necessarily reflect the views of MathWorks, Inc.}
\author{Ertuğrul Keçeci and Tufan Kumbasar}% <-this % stops a space
\thanks{Ertuğrul Keçeci and Tufan Kumbasar are with AI and Intelligent Systems Laboratory, Istanbul Technical University, 34469, Istanbul, Türkiye  {\tt\small kececie@itu.edu.tr, kumbasart@itu.edu.tr}}%
}

\begin{document}

\maketitle
\thispagestyle{empty}
\pagestyle{empty}

%%%%%%%%%%%%%%%%%%%%%%%%%%%%%%%%%%%%%%%%%%%%%%%%%%%%%%%%%%%%%%%%%%%%%%%%%%%%%%%%
\begin{abstract}
Recent advances in Deep Learning (DL) have strengthened data-driven System Identification (SysID), with Neural and Fuzzy Ordinary Differential Equation (NODE/FODE) models achieving high accuracy in nonlinear dynamic modeling. Yet, system states in these frameworks are often reconstructed without clear physical meaning, and input contributions to the state derivatives remain difficult to interpret. To address these limitations, we propose Explainable FODE (xFODE), an interpretable SysID framework with integrated DL-based training. In xFODE, we define states in an incremental form to provide them with physical meanings. We employ fuzzy additive models to approximate the state derivative, thereby enhancing interpretability per input. To provide further interpretability, Partitioning Strategies (PSs) are developed, enabling the training of fuzzy additive models with explainability. By structuring the antecedent space during training so that only two consecutive rules are activated for any given input, PSs not only yield lower complexity for local inference but also enhance the interpretability of the antecedent space. To train xFODE, we present a DL framework with parameterized membership function learning that supports end-to-end optimization. Across benchmark SysID datasets, xFODE matches the accuracy of NODE, FODE, and NLARX models while providing interpretable insights. 
\end{abstract}

%%%%%%%%%%%%%%%%%%%%%%%%%%%%%%%%%%%%%%%%%%%%%%%%%%%%%%%%%%%%%%%%%%%%%%%%%%%%%%%%
\section{Introduction}

Deep Learning (DL) has been increasingly applied for data-driven System Identification (SysID) to learn complex system dynamics \cite{pillonetto2025deep, dai2024deep, di2024stable, rnn_sysid}. Recent efforts include probabilistic deep models \cite{benavoli2025dynogp}, physics-informed architectures \cite{yu2024physics}, and hybrid gray–box designs \cite{de2023hybrid}. Among these, Neural Ordinary Differential Equations (NODEs) have emerged as a powerful framework that models state derivatives with Neural Networks (NNs) \cite{rahman2022neural}, and have been extended to improve training efficiency and interpretability \cite{node_sysid, rahman2022neural, bottcher2023gradient}.

Building on NODEs, Fuzzy Ordinary Differential Equations (FODEs) replace NNs with Fuzzy Logic Systems (FLSs) to provide interpretable system dynamics \cite{guven2025fuzzy}. Yet, the design in \cite{guven2025fuzzy} defines rules over high-dimensional antecedent spaces with unrestricted interactions, which limits its interpretability. Furthermore, since no explicit Partitioning Strategy (PS) is applied to the antecedent space, the learned Gaussian Membership Functions (MFs) can exhibit substantial overlap between neighboring MFs, reducing transparency. Thus, the interpretability of FLSs is underutilized, and FODEs become black-box models, like NODEs.

\begin{figure}[t] 
\includegraphics[width=\linewidth]{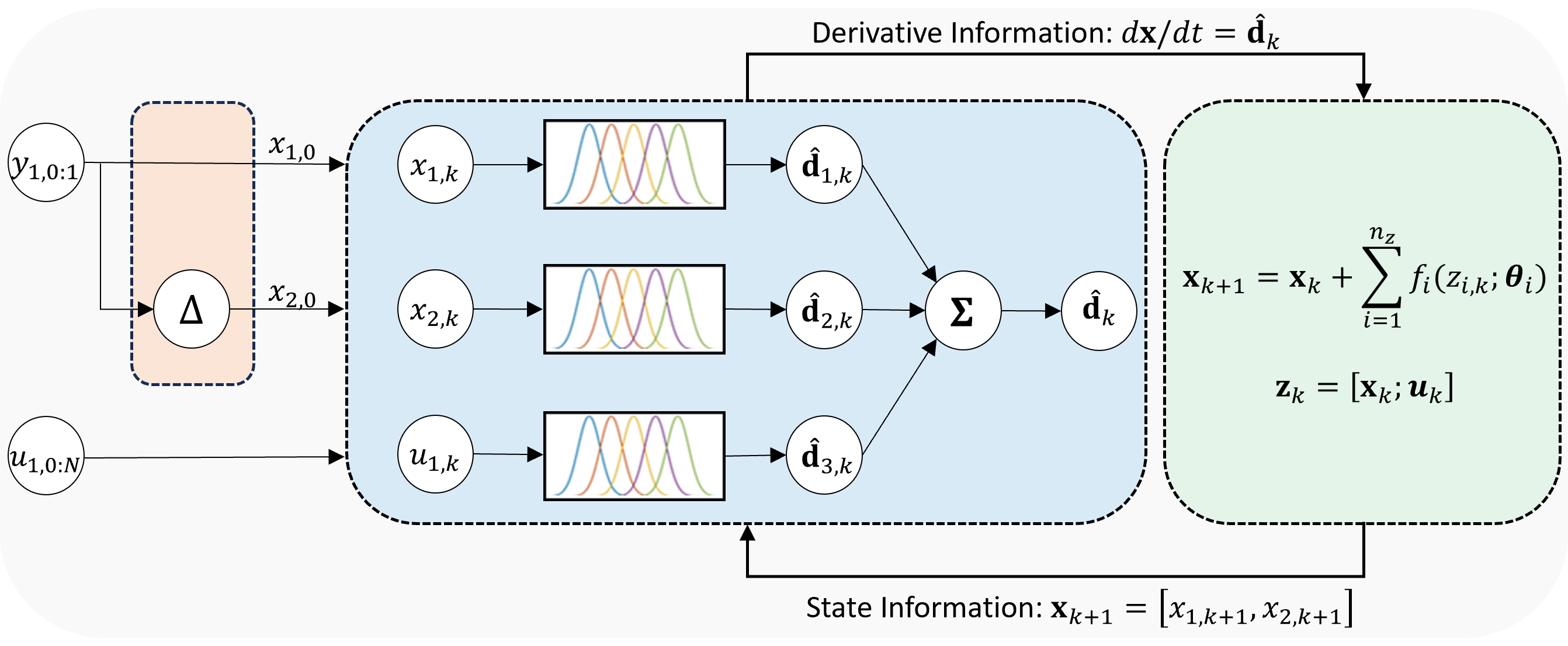}
\caption{Illustration of xFODE inference in a SISO setup with two states for clarity: Initial states are obtained from the \textcolor{paleorange}{State Representation} block, defined incrementally as $\mathbf{x}_k = [y_k, \Delta y_k]$. These states, together with the input, are fed to the \textcolor{skyblue}{FAME}, to generate the state derivatives. The derivatives are then integrated in the \textcolor{palegreen}{ODE} block to compute the next states.}
    \label{fig:xfode}
\end{figure}

Although NODEs and FODEs can accurately capture nonlinear system dynamics, they remain challenging to interpret. System states are often not directly measurable and must be estimated or reconstructed \cite{pillonetto2025deep, dai2024deep, di2024stable}, which can compromise their physical meanings. Moreover, both frameworks provide limited transparency regarding how each dimension of a high-dimensional state representation affects the system dynamics. Additive modeling structures \cite{agarwal2021neural,mariotti2023exploring,yang2021gami} provide a potential solution, as they can decompose system dynamics into dimension-wise contributions while preserving expressive power. Quite recently, Fuzzy Additive Models (FAMs) have been proposed to enhance interpretability on tabular data \cite{gokmen2025fame}, yet their application to SysID remains unexplored.

This paper proposes \emph{Explainable FODE} (xFODE), an interpretable data-driven SysID model with integrated DL based training. In xFODE, states are constructed from measurable outputs in an incremental form, preserving their physical meaning. To capture input-wise contributions to state dynamics, xFODE employs FAMs to approximate the state derivatives. Further, we introduce PSs for training of FAM, effectively transforming it into a FAM with Explainability (FAME) \cite{gokmen2025fame}. The PSs sculpt the antecedent space during training so that only two consecutive rules are active for any input, providing reduced local inference complexity and direct interpretability of the antecedent space. To train xFODE, we develop a DL framework using parameterization techniques that allow learning of MFs for each PS. To assess xFODE, we compare it with NODE, FODE, and NonLinear AutoRegressive network with eXogenous inputs (NLARX) models on SysID datasets. The results show that xFODE achieves interpretable modeling with accuracy on par with NODE, FODE, and NLARX.

\section{Neural and Fuzzy ODE Networks}
This section provides a concise overview of NODE and FODE and their design challenges. Let us define a non-autonomous partially observable nonlinear system as:
\begin{equation}
\label{dynamics}
\begin{aligned}
    \dot{\mathbf{x}}(t) &= d(\mathbf{x}(t),\mathbf{u}(t)), \ \mathbf{x}(t_0) = \mathbf{x}_0 \\
    \mathbf{y}(t) &= h(\mathbf{x}(t))
\end{aligned}
\end{equation}
where $\mathbf{x}(t) \in \mathbb{R}^{n_x}$ denotes the states, $\mathbf{u}(t) \in \mathbb{R}^{n_u}$ the inputs, and $\mathbf{y}(t) \in \mathbb{R}^{n_y}$ the outputs. $d(\cdot)$ describes the nonlinear model, while $h(\cdot)$ defines the output mapping. 

\subsection{Dynamic Inferences of NODE and FODE}

To represent the dynamics in \eqref{dynamics}, it has been shown that NODEs \cite{chen2018neural} and FODEs \cite{guven2025fuzzy} are efficient structures. Both models are defined with a parameterized vector field $d^{\mathrm{M}}(\cdot)$ ($\mathrm{M} \in \{\mathrm{NN},\,\mathrm{FLS}\}$) as follows: 
\begin{equation} \label{neural_dynamics}
    \dot{\mathbf{x}}(t) = d^\mathrm{M}\left(\mathbf{x}(t), \mathbf{u}(t); {\theta} \right)
\end{equation}
where ${\theta}$ represents the Learnable Parameters (LPs). Given the initial condition $\mathbf{x}(t_0)$, the state trajectory over the interval $[t_0,\, t_1]$ is then expressed as
\begin{equation}
\label{xtraj}
    \mathbf{x}(t_1) = \mathbf{x}(t_0) + \int_{t_0}^{t_1} d^\mathrm{M}(\mathbf{x}(\tau),\, \mathbf{u}(\tau); \theta)\, d\tau.
\end{equation}
If the Euler integration method is used, then the discrete-time representation of \eqref{dynamics} is as follows: 
\begin{equation}
\label{discnonlin}
\begin{aligned}
    \mathbf{x}_{k+1} &= \mathbf{x}_k + d^\mathrm{M}(\mathbf{z}_k) \\
    \mathbf{y}_k &= h(\mathbf{x}_k),
\end{aligned}
\end{equation}
where the input $\mathbf{z}_k = [\mathbf{x}_k;\ \mathbf{u}_k] \in \mathbb{R}^{n_z}$ is defined as $\mathbf{z}_k = [z_{1,k}, z_{2,k}, \dots, z_{n_z,k}]$ with $n_z = n_x + n_u$.

\subsection{Challenges in NODE and FODE Modeling}

Although NODEs and FODEs provide flexible dynamic representations, two fundamental challenges limit their interpretability and practical utility. 
\begin{itemize}
\item \textbf{State Representation:}
The state vector $\mathbf{x}_k$ is generally \emph{not directly measurable}. Thus, it must be estimated or defined \cite{pillonetto2025deep}, which might destroy their physical meaning and render the learned dynamics difficult to interpret. To preserve interpretability, the states can be defined directly from the outputs $\mathbf{y}_k = [y_{1,k},\, y_{2,k},\, \dots,\, y_{n_y,k}]^T \in \mathbb{R}^{n_y}$ rather than latent or estimated states.

\item \textbf{Lack of Interpretability of $d^{\mathrm{M}}$:}
Regardless of whether $d^{\mathrm{M}}(\mathbf{z}_k)$ is implemented via a NN or an FLS, it functions as a high-dimensional black-box mapping. In the case of NODEs, the NN captures complex nonlinearities but provides no direct insight into the contribution of individual inputs. For FODEs, although fuzzy rules are employed, the high-dimensional antecedent spaces and unrestricted rule interactions often yield entangled dynamics that are difficult to interpret. Thus, the influence of each input $z_{i,k}$ on the state derivative remains opaque.
    
\end{itemize}
As a result, these models cannot provide input-wise attribution, model transparency, or component-level interpretability.

\section{xFODE Framework}

The limitations of NODE and FODE motivate the need for more transparent models. As shown in Fig.~\ref{fig:xfode}, the proposed xFODE aims to preserve the expressive power of NODE/FODE modeling while enabling interpretability. The following subsections detail its key components.

\subsection{Interpretable State Representation}\label{staterep}

A standard approach for constructing the state vector $\mathbf{x}_k$ is the \emph{lagged form}, which stacks a sequence of past outputs:
\begin{flalign} \label{lagged}
  \text{SR1: } && \mathbf{x}_k = [\,\mathbf{y}_k,\, \mathbf{y}_{k-1},\, \dots,\, \mathbf{y}_{k-m}\,]^T, &&
\end{flalign}
where $m$ denotes the number of lags. While this formulation captures temporal dependencies, each component represents a past output, making it difficult to directly link elements of $\mathbf{x}_k$ to the system’s dynamic evolution.

To improve interpretability, we suggest defining $\mathbf{x}_k$ in the \emph{incremental form}, in which the state vector is constructed using discrete differences:
\begin{flalign}
\label{incremental}
   \text{SR2: }&& \mathbf{x}_k = [\,\mathbf{y_k},\,\Delta \mathbf{y}_k,\, \dots,\, \Delta^m \mathbf{y}_k\,]^T&&
\end{flalign}
with $m$ denoting the number of difference orders and $\Delta^j \mathbf{y}_k = \Delta^{j-1}\mathbf{y}_k - \Delta^{j-1}\mathbf{y}_{k-1}, j = 1, \dots, m$. This representation emphasizes temporal changes, with each term corresponding to a physically meaningful quantity, such as velocity or acceleration of the outputs. 

\subsection{Additive ODE Dynamics}

Modeling the state derivative as a single high-dimensional mapping can obscure input contributions. xFODE instead represents the derivative additively:
\begin{equation}
\label{xfode}
    \mathbf{x}_{k+1} = \mathbf{x}_k + \sum_{i=1}^{n_z} f_i(z_{i,k}; \boldsymbol{\theta}_i),
\end{equation}
where $f_i: \mathbb{R} \to \mathbb{R}^{n_x}$ is a single-input mapping for $z_{i,k}$. Each mapping produces $n_x$ outputs, one for each state dimension, 
\begin{equation}
    \hat{\mathbf{d}}_{i,k} = f_i(z_{i,k}; \boldsymbol{\theta}_i),
\end{equation}
where $\hat{\mathbf{d}}_{i,k} \in \mathbb{R}^{n_x}$ represents the contribution of input $z_{i,k}$ to each component of the derivative-like update of the state vector. This additive decomposition enables input-wise interpretability while retaining the capacity to model complex dynamics. 

\begin{figure*}[t]
        \centering
        \subfigure[Learned GaussMF if there is no PS employed]
        {
        \includegraphics[width=0.35\textwidth,height=0.21\textheight,keepaspectratio]{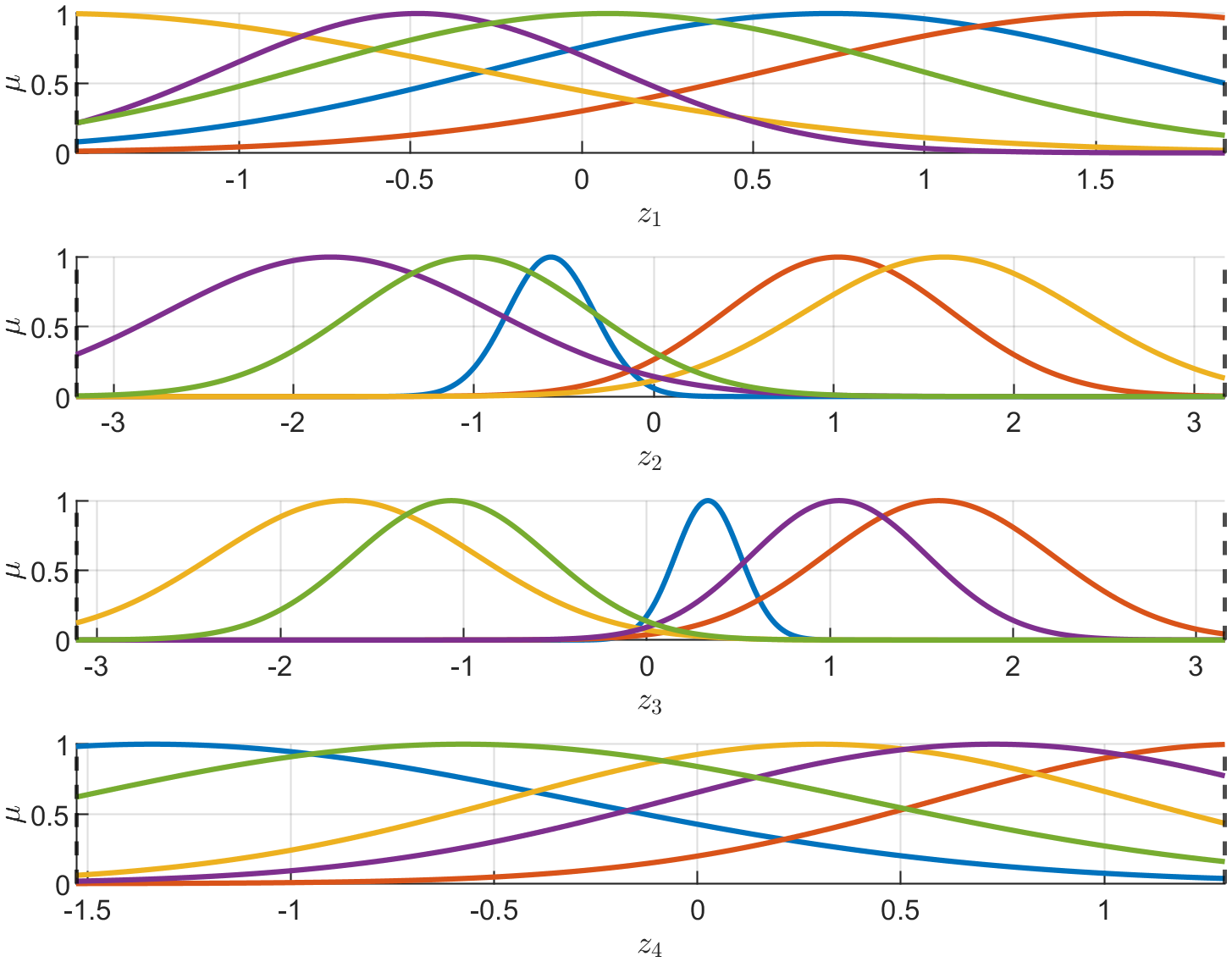}\label{fig:gaussmf}

        }
        \subfigure[Learned TriMFs via PS1]
        {
        \includegraphics[width=0.35\textwidth,height=0.21\textheight,keepaspectratio]{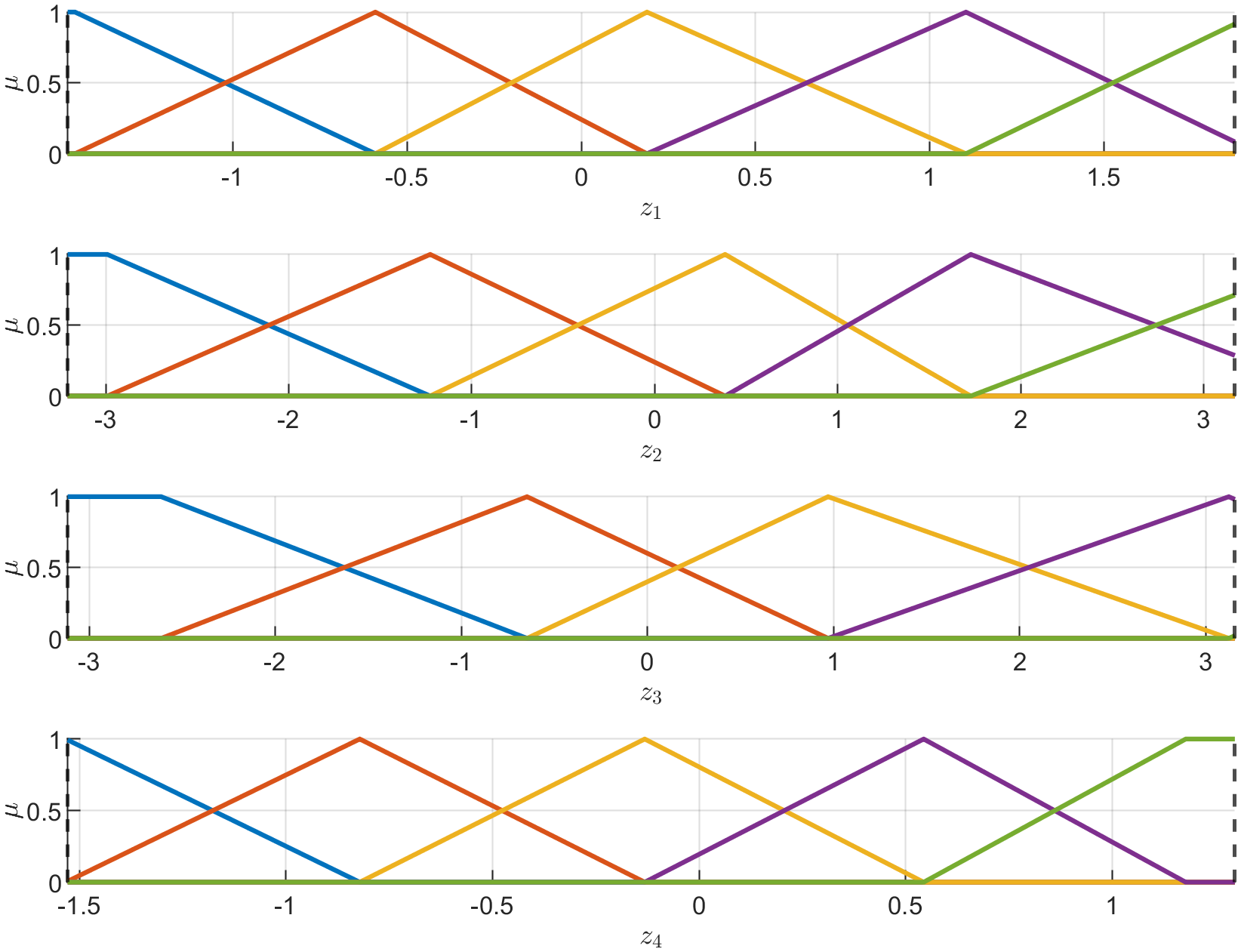}\label{fig:trimf}
        
        }

        \subfigure[Learned Gauss2MF via PS2]
        {
        \includegraphics[width=0.35\textwidth,height=0.21\textheight,keepaspectratio]{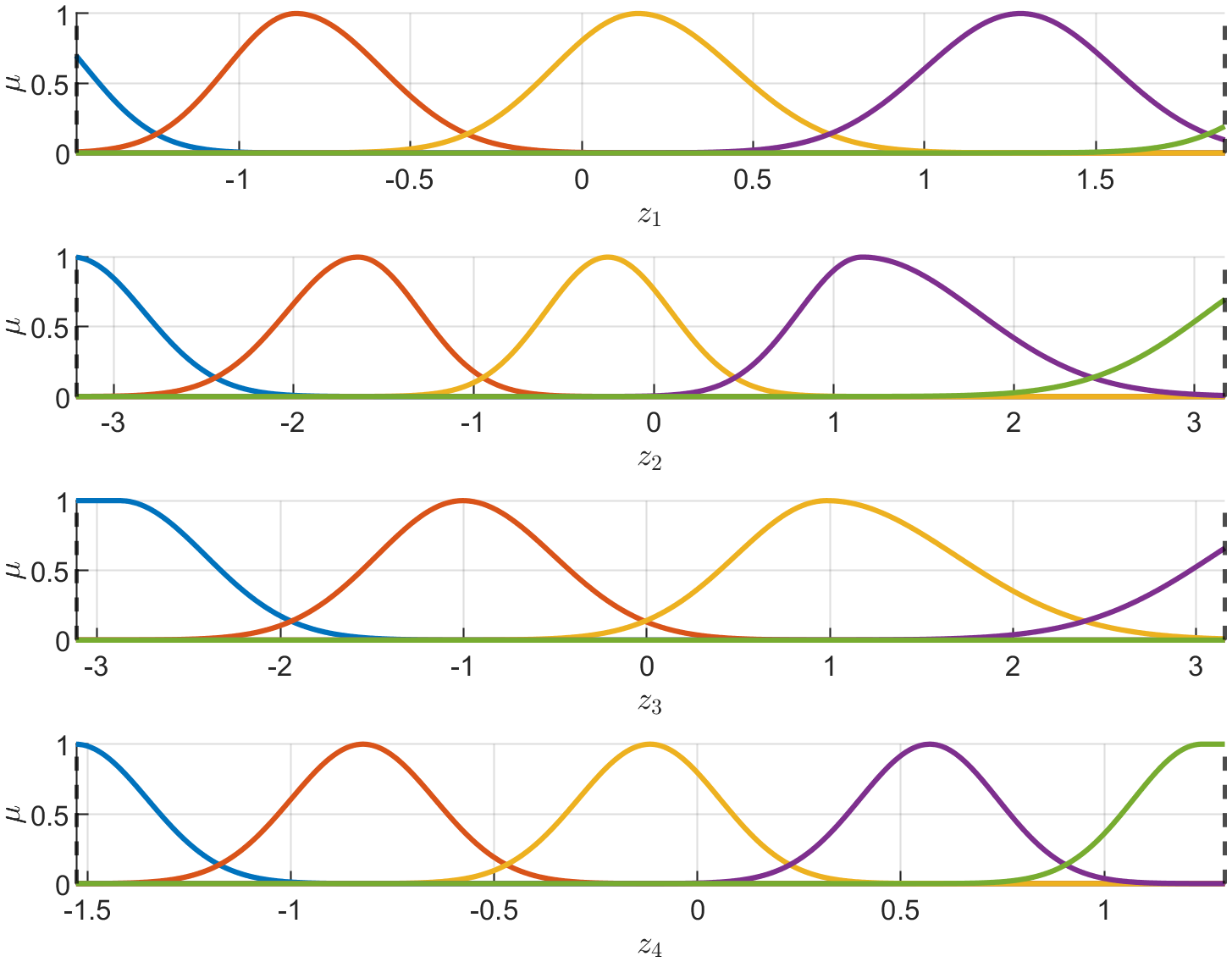}\label{fig:gauss2mf-1}
        }
        \subfigure[Learned Gauss2MF via PS3]
        {
        \includegraphics[width=0.35\textwidth,height=0.21\textheight,keepaspectratio]{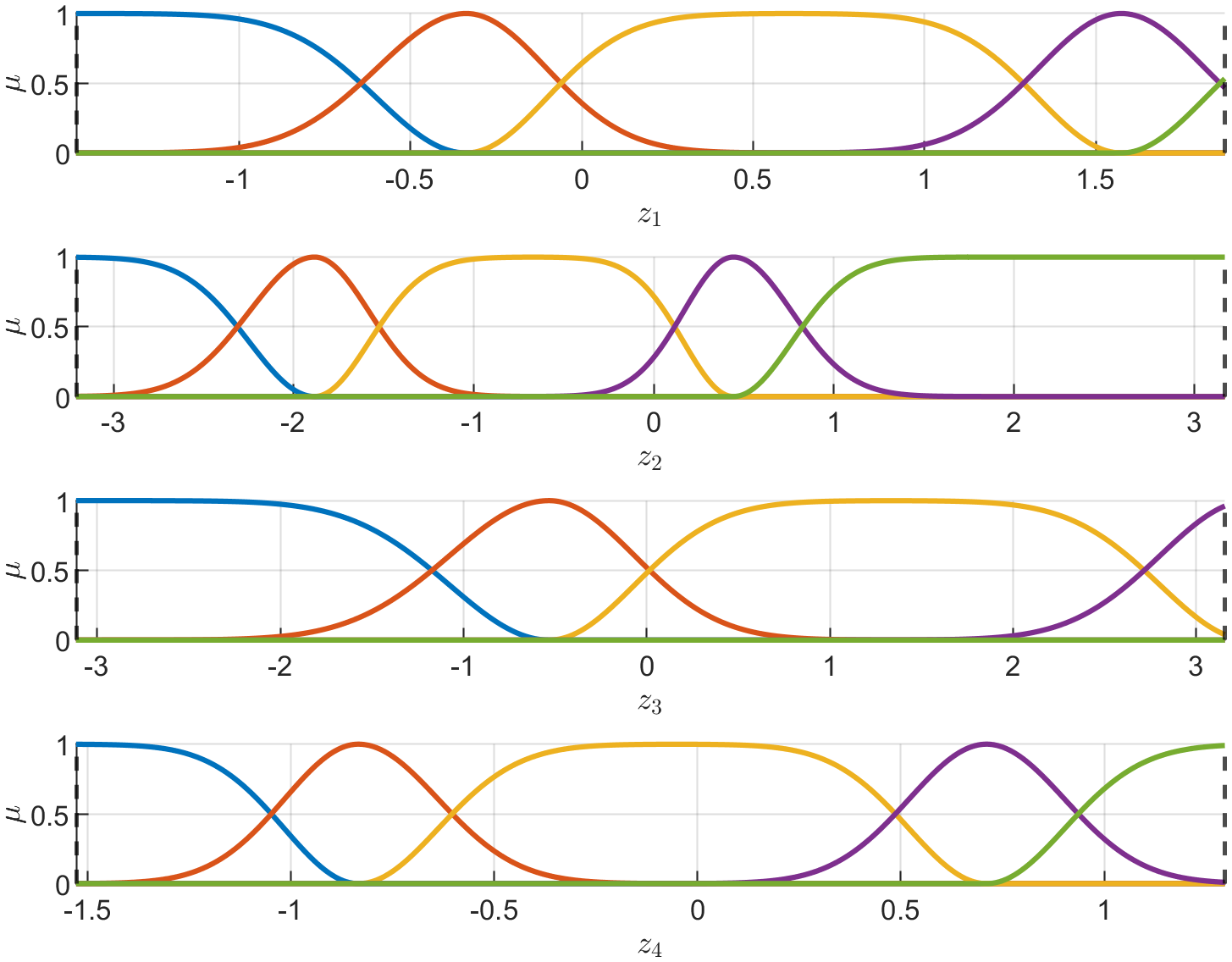}\label{fig:gauss2mf-2}
        
        }
        \caption{Visualization of learned MFs on the Two-Tank dataset (single seed). States are defined with the incremental form (SR2), thus $\mathbf{z} = [\mathbf{y}; \Delta\mathbf{y}; \Delta^2\mathbf{y};\mathbf{u}]$. Each combined input dimension $z_i$ is interpretable: \(z_1\) output position, \(z_2\) output velocity, \(z_3\) output acceleration, and \(z_4\) control input.}
        \label{fig:MFs}
\end{figure*}

\subsection{Interpretable Mapping Design} \label{antdesign}

In xFODE, each mapping $f_i(.)$ is implemented as a single-input FLS, producing $n_x$ outputs that correspond to contributions to each state dimension $\mathbf{x}_k$. For notational simplicity, we drop the indices $i$ and $k$ throughout this subsection.

Each FLS consists of $P$ rules ($p=1,2,\dots,P$) defined as
\begin{equation}
    R_p: \text{If $z$ is $A_p$ Then $\hat{\mathbf{d}}$ is $\mathbf{d}_p$},
\end{equation}
with $\mathbf{d}_p = [d_{p,1}, d_{p,2}, \dots, d_{p,n_x}]^T$ and the  consequents are defined as $d_{p,o} = a_p^o z + a_{p,0}^o, o=1,2,\dots,n_x$. The FLS mapping can be defined in a compact form as follows:
\begin{equation}
\label{flsinf}
    f(z;\boldsymbol{\theta}) = \frac{\sum_{p=1}^P\mu_p(z)\mathbf{d}_p}{\sum_{p=1}^P\mu_p(z)},
\end{equation}
where $\mu_{p}(z)$ is the membership grade of the fuzzy set $A_p$. This formulation allows all output dimensions to share the same antecedent space while maintaining independent consequents for each state dimension, enabling transparent input-wise contribution to the state derivative.

A common choice for defining the input domain is using Gaussian MFs (GaussMFs) to capture nonlinearities:
\begin{equation}
\label{gaussmf}
    \mu_{p}(z) = \text{exp}(-(z-c_p)^2 / 2(\sigma_p)^2) 
\end{equation}
where $c_p$ and $\sigma_p$ are the center and standard deviation, respectively. While GaussMFs provide smooth transitions, their overlap can cause multiple rules to be simultaneously fired, reducing interpretability, as illustrated in Fig. \ref{fig:gaussmf}. 

To address this, we propose PSs for sculpting the antecedent space so that, for any input, only two consecutive rules are active. Thus, the mapping in \eqref{flsinf} simplifies to  
\begin{equation}
\label{flsinf2}
    f(z';\boldsymbol{\theta}) = \frac{\sum_{p=p^*}^{p^*+1} \mu_p(z') \mathbf{d}_p}{\sum_{p=p^*}^{p^*+1} \mu_p(z')}.
\end{equation}
for \(z' \in [c_{p^*},\, c_{p^*+1}]\). This ensures a simple and interpretable mapping while maintaining computational efficiency. 

\subsubsection{PS1-Partitioning with Triangular MFs (TriMFs)} TriMFs offer a straightforward way to partition the input domain. A TriMF is defined as follows:
\begin{equation}\label{trimf}
\mu_p(z) = \max\Big(0, \min\Big(\frac{z-l_p}{c_p-l_p}, \frac{r_p-z}{r_p-c_p}\Big)\Big)
\end{equation}
where $l_p$, $c_p$, and $r_p$ denote the left endpoint, center, and right endpoint of the $p^{\text{th}}$ MF. 

To maintain interpretable partitioning, the following coupling relationships are defined between consecutive MFs:
\begin{align}
\Delta_p^l = c_p - l_p, \quad
\Delta_p^r = r_p - c_p,
\\
c_{p+1} = c_p + \Delta_p^r, \quad
\Delta_{p+1}^l = \Delta_p^r.
\end{align}
This coupling ensures that, for any input $z'$, only two consecutive MFs are activated as shown in Fig. \ref{fig:trimf}. 

\subsubsection{PS2-Partitioning with Two-Sided Gaussian MFs (Gauss2MFs)}
To provide greater flexibility and smoothness while preserving interpretability, we can use Gauss2MFs:
\begin{equation}
\label{gauss2mf}
\mu_{p}(z)=\left\{\begin{array}{l}
\exp \left(-\left(z-c_p\right)^2 / 2\left(\sigma_p^l\right)^2\right), \text { if } z \leq c_p \\
\exp \left(-\left(z-c_p\right)^2 / 2\left(\sigma_p^r\right)^2\right), \text { if } z>c_p
\end{array}\right.
\end{equation}
where $\sigma^l_p$ and $\sigma^r_p$ are the left and right standard deviations, respectively. 

To ensure interpretability and smooth overlap between adjacent MFs, we parameterize the centers as: 
\begin{equation}
    \label{centers}
        c_{p+1} = c_p + 4\sigma^r_p
    \end{equation}
and the standard deviations as: 
\begin{equation}
    \label{sigmas}
        \sigma^r_p = \sigma^l_{p+1}
\end{equation}
This parameterization ensures an interpretable partition of the input domain as shown in Fig. \ref{fig:gauss2mf-1}. Note that we assume $\mu_{p}(z') \approx 0$ when $|z'|>4\sigma^r_p$ \cite{gokmen2025fame}.  
% for an input $z'$, only two consecutive rules, $\mu_{p}(z')$ and $\mu_{{p+1}}(z')$, are activated since $\mu_{p}(z') \approx 0$ when $|z'|>4\sigma^r_p$. Thus, we can obtain the mapping in \eqref{flsinf2}. 

\subsubsection{PS3-Partitioning with Complementary Gauss2MFs} 
In PS2, while Gauss2MFs offer asymmetric flexibility, the $4\sigma$ center spacing in \eqref{centers} may result in weak overlap between adjacent MFs. Thus, to increase the overlap, we construct odd-indexed MFs as local complements of neighboring even-indexed MFs. Let even indexed MFs \(A_{2q}\) (\(q=1,2,\dots, \left\lceil (P-3)/2 \right\rceil\)) be defined by \eqref{gauss2mf}. The midpoint between consecutive even-indexed MFs is:
\begin{equation}
\bar{c}_{2q+1} = ({c_{2q} + c_{2(q+1)}})/{2}.
\end{equation}
The odd indexed MFs \(A_{2q+1}\) are then defined as
\begin{equation}
\mu_{2q+1}(z) =
\begin{cases}
0, & z < c_{2q},\\
1-\mu_{2q}(z), & c_{2q} \le z \le \bar{c}_{2q+1},\\
1-\mu_{2(q+1)}(z), & \bar{c}_{2q+1} \le z \le c_{2(q+1)},\\
0, & z > c_{2(q+1)},
\end{cases}
\end{equation}
with \(\mu_1(z) = 1-\mu_2(z), z < c_2\) and when $P$ is odd, \(\mu_P(z) = 1-\mu_{P-1}(z), z > c_{P-1}\). This MF design guarantees that, for any input, only two MFs are active, as shown in Fig. \ref{fig:gauss2mf-2}. 

\section{DL Framework for xFODE}

In this section, we present a DL framework for learning xFODE. Algorithm \ref{alg:alg1}\footnote[1]{MATLAB implementation. [Online]. Available: \newline \url{https://github.com/ertugrulkececi/xfode}} outlines the overall training process for xFODE, while Algorithm \ref{alg:alg2} details the training inference applied to each trajectory within a mini-batch. Here, we consider a dataset $D = \{\mathbf{u}_k, \mathbf{y}_k\}_{k=1}^K$. The state vector \(\mathbf{x}_k\) is constructed from \(\mathbf{y}_k\) using either the lagged or incremental SR. Then, we define the input–state trajectories for training. 
\begin{equation}
S=\left[\left(\mathbf{u}_0^{\{j\}}, \mathbf{x}_0^{\{j\}}\right), \ldots,\left(\mathbf{u}_N^{\{j\}}, \mathbf{x}_N^{\{j\}}\right)\right], j \in[1, \ldots, B]
\end{equation}
Here, $N$ is the roll-out while $B$ is the number of trajectories.

xFODE learns the underlying system dynamics by minimizing an $L_1$ loss function over $S$:
\begin{equation}
\label{l1loss}
    \min_{\boldsymbol{\theta \in \mathcal{C}}} L_1 = \frac{1}{B}\sum_{j=1}^B \sum_{k=1}^N |\mathbf{x}_k^{\{j\}} - \hat{\mathbf{x}}_k^{\{j\}}|
\end{equation}
where $\hat{\mathbf{x}}_k^{\{j\}}$ represents the predicted states by xFODE:
\begin{equation}
    \hat{\mathbf{x}}_{k} = \hat{\mathbf{x}}_{k-1} + \sum_{i=1}^{n_z} f_i(\hat{z}_{i,k-1}; \boldsymbol{\theta}_i).
\end{equation}
The complete set of LPs is defined as $\boldsymbol{\theta} = \{\boldsymbol{\theta}_i\}^{n_z}$ where each $\boldsymbol{\theta}_i$ is parameterized according PS:
\begin{itemize}
    \item For PS1: $\boldsymbol{\theta}_i = \{c_1,\Delta_1^l,\boldsymbol{\Delta}^r, \mathbf{d}\}$ where $c_1$ and $\Delta_1^l$ are scalar, $\boldsymbol{\Delta}^r \in \mathbb{R}^{P \times 1}$, and $\mathbf{d} \in \mathbb{R}^{2Pn_x \times 1}$
    \item For PS2/ PS3: $\boldsymbol{\theta}_i = \{c_1,\sigma_1^l,\boldsymbol{\sigma}^r, \mathbf{d}\}$ where $c_1$ and $\sigma_1^l$ $\sigma$ are scalar, $\boldsymbol{\sigma}^r \in \mathbb{R}^{P \times 1}$, and $\mathbf{d} \in \mathbb{R}^{2Pn_x \times1}$
\end{itemize}

\begin{algorithm}[t]
\caption{Training steps of xFODE} \label{alg:alg1}
\begin{algorithmic}[1]
\State \textbf{Input:} $D = \{\mathbf{u}_k, \mathbf{y}_k\}_{k=1}^K$, dataset
\State $B$, number of measured trajectories
\State $N$, prediction horizon
\State $P$, number of rules
\State $mbs$, mini-batch size
\State $E$, number of epochs
\State $m$, number of lags/ difference orders in \eqref{lagged}/\eqref{incremental}
\State \textbf{Output:} LP set ${\theta}$
\State Initialize ${\theta}$;
\State Construct $\mathbf{x}_k$ via \eqref{lagged}/\eqref{incremental};
\State Build $S = (\mathbf{u}^{\{j\}}_{1:N},\mathbf{x}^{\{j\}}_{1:N})^{B}_{j=1}$;
\For{$e = 1 \text{ to } E$}
\For{\textbf{each } $mbs$ in $B$} 
    \State $\text{Select a mini-batch } [ \mathbf{u}^{\{j\}}_{0:N},\mathbf{x}^{\{j\}}_{0:N}]^{mbs}_{j=1}$
    \State $\hat{\mathbf{x}}^{\{j\}}_{1:N} \leftarrow \text{xFODE}([\mathbf{x}^{\{j\}}_{0}, \mathbf{u}^{\{j\}}_{0:N}]; {{\theta}})$
    \State Compute $L_1$
    \State Compute the gradient ${\partial L_1}/{\partial {\theta}} $ via AD 
    \State Update ${\theta}$ via a DL optimizer, e.g., Adam 
\EndFor
\EndFor
\State ${\theta}^* = {\arg \min }(L_1)$
\State \textbf{Return} $\theta = {\theta}^{*}$
\end{algorithmic}
\end{algorithm}

\begin{algorithm}[t]
\caption{Training inference of xFODE for a trajectory in a mini-batch} \label{alg:alg2}
\begin{algorithmic}[1]
\State \textbf{Input:} Initial state $\mathbf{x_0}$, input sequence $\mathbf{u}_{0:N}$, LP set ${\theta}$, prediction horizon $N$
\State \textbf{Output:} Predicted state trajectory $\hat{\mathbf{x}}_{1:N}$
\State Initialize $\hat{\mathbf{x}} \leftarrow \varnothing$, $\hat{\mathbf{x}}_0 = \mathbf{x}_0$
\For{$k = 0$ \textbf{to} $N-1$}
    \State $\mathbf{z_k} \leftarrow [\hat{\mathbf{x}}_{k}, \mathbf{u}_{k}]$
    \State $\hat{\mathbf{x}}_{k+1} \gets \hat{\mathbf{x}}_{k} + \sum_{i=1}^{n_z} f_i(\hat{z}_{i,k}; \boldsymbol{\theta}_i)$
    \State $\hat{\mathbf{x}} \leftarrow \hat{\mathbf{x}} \cup \hat{\mathbf{x}}_{k+1}$ 
    \State $\hat{\mathbf{x}}_{k} \leftarrow \hat{\mathbf{x}}_{k+1}$
\EndFor
\State \textbf{Return} $\hat{\mathbf{x}}$
\end{algorithmic}
\end{algorithm}

During training, it is necessary to enforce $\sigma_p > 0$ or $\Delta_p > 0$, $\forall p$, depending on the shape of MF \cite{beke2022more}. Since DL optimizers operate in unconstrained parameter spaces, we reparameterize the MF spreads to enable unconstrained optimization. For TriMF, we define $\Delta_p' \in (-\infty, \infty)$:
    \begin{equation}
        \Delta_p = \log(1 + e^{\Delta_p'})
    \end{equation}
and $\sigma_p' \in (-\infty, \infty)$ for Gauss2MF as: 
    \begin{equation}
        \sigma_p = \log(1 + e^{\sigma_p'})
\end{equation}

\section{Performance Analysis}
We evaluate xFODE, FODE, and NODE on five benchmark SysID datasets summarized in Table~\ref{tab:datasets}, with all datasets normalized. Experiments are conducted in a MATLAB environment and repeated over 20 independent runs for statistical reliability. 
The analyzed model configurations are:
\begin{itemize}
    \item \textbf{NODE:} $d^{NN}(\cdot)$ is a neural network with two hidden layers of 128 units each and \textit{tanh} activations.
    \item \textbf{FODE:} $d^{FLS}(\cdot)$ is a multi-input FLS with $P=5$ rules, using GaussMFs in the antecedents without a PS.
    \item \textbf{xFODE:} Each FLS $f_i(\cdot)$ in \eqref{xfode} uses $P=5$ rules with antecedent sculpting strategies PS1--PS3 (Section~\ref{antdesign}).
    \item \textbf{Additive FODE (AFODE):} To evaluate the effect of PSs, we include an AFODE variant using GaussMFs without a PS, with $P=5$ rules.
\end{itemize}
For all models, the state vector $\mathbf{x}_k$ is constructed using either SR1 or SR2. The SR hyperparameter $m$ is selected from the range $[0,5]$ via cross-validation on NODE and set to $m=2$ for the Two-Tank, Hair Dryer, and MR Damper; and $m=1$ for the Steam Engine and EV Battery datasets. The roll-out is set to $N=20$ across all datasets. 

To test their SysID performances, all models are evaluated using multi-step prediction tasks (i.e., simulation mode). We report the mean RMSE along with its ±1 standard error in Table \ref{tab:perf_models_rows_ieee_g0_ev}, as well as \#LP for each model. Moreover, for benchmarking, NLARX models were trained using MATLAB's built-in  \texttt{nlarx} SysID function with Sigmoid Network (SN), Tree Ensemble (TE), Support Vector Machine (SVM), and NN. We reported their regressor definition and RMSE values in Table \ref{tab:nlarx}. The results indicate that:

\begin{table}[t]
    \centering
    \caption{Datasets summary}
    \label{tab:datasets}
    \begin{tabular}{|l|l|l|l|l|}
        \hline
        \textbf{Dataset} & $n_u$ & $n_y$ & \textbf{Training samples} & \textbf{Testing samples} \\
        \hline
        Two-Tank     & 1 & 1 & 1500 & 1500 \\
        Hair Dryer   & 1 & 1 &   500 &   500 \\
        MR Damper    & 1 & 1 & 3000 &   499 \\
        Steam Engine & 2 & 2 &   250 &   201 \\
        EV Battery   & 2 & 1 & 15001 & 14351 \\
        \hline
    \end{tabular}
\end{table}

\begin{figure*}[t]
        \centering
        \subfigure[Two-Tank dataset]
        {
        \includegraphics[width=\linewidth,height=0.12\textheight,keepaspectratio]{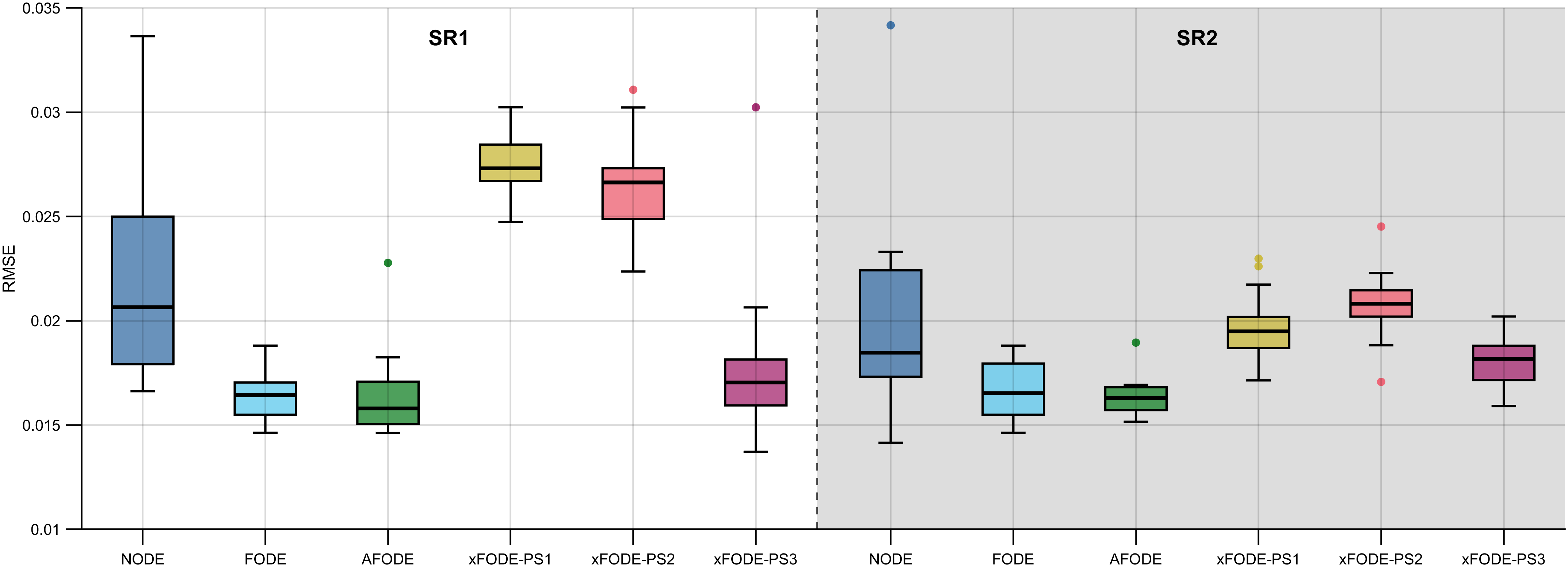}\label{fig:ttmatlab}

        }
       \subfigure[MR Damper dataset]
        {
        \includegraphics[width=\linewidth,height=0.12\textheight,keepaspectratio]{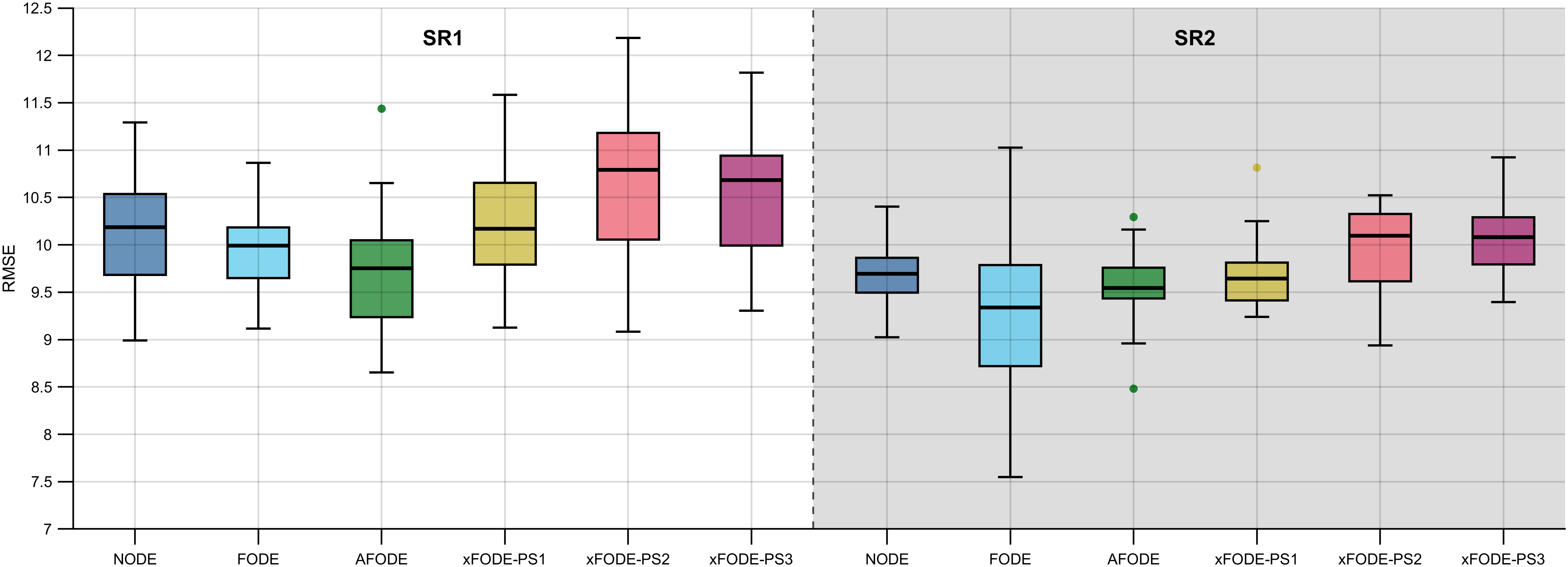}\label{fig:mrdamper}

        }

        \caption{RMSE boxplots of trained models on (a) Two-Tank and (b) MR-Damper datasets; SR1 with the white, SR2 with the gray background.}
        \label{fig:boxplots}
\end{figure*}

\begin{itemize}
    \item Using SR2 to define states consistently yields lower RMSE and variance than SR1 for all models, as also shown from the box plot in Fig.~\ref{fig:boxplots}. 
    \item xFODE achieves lower RMSE than NODE on the Two-Tank and Steam Engine datasets and outperforms FODE on the Hair Dryer, EV Battery, and Steam Engine datasets. While NODE and FODE perform slightly better on some datasets, xFODE’s improved interpretability comes with minimal performance cost and a comparable \#LP to FODE, markedly fewer than NODE.
    \item xFODE yields slightly higher RMSE than AFODE across all datasets. However, unlike AFODE, where multiple rules can be active simultaneously (Fig. \ref{fig:gaussmf}), xFODE learns rules such that only two consecutive rules are activated at any time (Figs. \ref{fig:trimf}--\ref{fig:gauss2mf-2}).
    \item xFODE attains its best SysID performance with different PSs depending on the dataset. These findings indicate that the best PS is dataset dependent. 
    \item xFODE outperforms TE and SVM on most datasets. While SN and NN methods enable NLARX models to achieve lower RMSE on the Hair Dryer, MR Damper, and EV Battery datasets, these models often require large regressor structures and lack interpretability.
\end{itemize}

In summary, the proposed xFODE framework effectively enhances interpretability through structured partitioning of the antecedent space. This key improvement is illustrated in Fig. \ref{fig:MFs}, where the antecedent spaces of the learned models can be 
expressed using intuitive linguistic terms such as Negative (N), Negative-Medium (NM), Zero (Z), Positive-Medium (PM), and Positive (P). Since MFs learned by xFODE are well-separated in contrast to AFODE, it enhances the human interpretability of the fuzzy rule base. xFODE delivers this with a low \#LP and performance on par with NODE/FODE, combined with the benefits of defining states within SR2.

\begin{table*}[t]
\centering
\caption{Testing performance of the NODE, FODE, and xFODE models: RMSE reported as mean $\pm$1 standard error over 20 experiments}
\label{tab:perf_models_rows_ieee_g0_ev}
\renewcommand{\arraystretch}{1.12}
\setlength{\tabcolsep}{2pt}
\begin{threeparttable}
\resizebox{\textwidth}{!}{%
\begin{tabular}{|l|c|c|c|c|c|c|c|c|c|c|}
\hline
\multirow{2}{*}{\textbf{Model}}
& \multicolumn{2}{c|}{\textbf{Two-Tank}}
& \multicolumn{2}{c|}{\textbf{Hair Dryer}}
& \multicolumn{2}{c|}{\textbf{MR Damper}}
& \multicolumn{2}{c|}{\textbf{EV Battery}}
& \multicolumn{2}{c|}{\textbf{Steam Engine}} \\
\cline{2-11}
& \#LP & RMSE & \#LP & RMSE & \#LP & RMSE & \#LP & RMSE & \#LP & RMSE \\ \hline

NODE-SR1 & 17539 & 0.0220 $\pm$ (0.0049) & 17539 & 0.1491 $\pm$ (0.0307) & 17539 & 10.1554 $\pm$ (0.6083) & 17410 & 0.1846 $\pm$ (0.0262) & 17924 & \begin{tabular}{@{}c@{}}$y_1$: 0.1151 $\pm$ (0.0051)\\$y_2$: 0.1011 $\pm$ (0.0051)\end{tabular} \\ \hline
NODE-SR2 & 17539 & 0.0197 $\pm$ (0.0043) & 17539 & {0.1173 $\pm$ (0.0038)} & 17539 & 9.6809 $\pm$ (0.3299) & 17410 & 0.1757 $\pm$ (0.0038) & 17924 & \begin{tabular}{@{}c@{}}$y_1$: 0.0909 $\pm$ (0.0078)\\$y_2$: 0.0873 $\pm$ (0.0056)\end{tabular} \\ \hline

FODE-SR1 & 115 & 0.0165 $\pm$ (0.0013) & 115 & 0.1638 $\pm$ (0.0154) & 115 & 9.9203 $\pm$ (0.4495) & 90 & 0.8918 $\pm$ (1.0916)* & 200 & \begin{tabular}{@{}c@{}}$y_1$: 0.0878 $\pm$ (0.0096)\\$y_2$: 0.1047 $\pm$ (0.0099)\end{tabular} \\ \hline
FODE-SR2 & 115 & 0.0166 $\pm$ (0.0014) & 115 & 0.1352 $\pm$ (0.0107) & 115 & {9.2819 $\pm$ (0.8725)} & 90 & 0.3814 $\pm$ (0.2994) & 200 & \begin{tabular}{@{}c@{}}$y_1$: 0.1014 $\pm$ (0.0171)\\$y_2$: 0.1129 $\pm$ (0.0184)\end{tabular} \\ \hline

AFODE-SR1 & 160 & 0.0164 $\pm$ (0.0018) & 160 & 0.1975 $\pm$ (0.0046) & 160 & 9.7185 $\pm$ (0.6903) & 120 & 0.1881 $\pm$ (0.0173) & 300 & \begin{tabular}{@{}c@{}}$y_1$: 0.1007 $\pm$ (0.0086)\\$y_2$: 0.1171 $\pm$ (0.0086)\end{tabular} \\ \hline
AFODE-SR2 & 160 & {0.0163 $\pm$ (0.0008)} & 160 & 0.1233 $\pm$ (0.0021) & 160 & 9.5588 $\pm$ (0.4112) & 120 & 0.1721 $\pm$ (0.0258) & 300 & \begin{tabular}{@{}c@{}}$y_1$: {0.0762 $\pm$ (0.0062)}\\$y_2$: {0.0713 $\pm$ (0.0049)}\end{tabular} \\ \hline

% PS1 is old PS3
xFODE-SR1-PS1 & 148 & 0.0275 $\pm$ (0.0014) & 148 & 0.1948 $\pm$ (0.0051) & 148 & 10.2756 $\pm$ (0.6852) & 108 & 0.2113 $\pm$ (0.0247) & 282 & \begin{tabular}{@{}c@{}}$y_1$: 0.1081 $\pm$ (0.0119)\\$y_2$: 0.1207 $\pm$ (0.0091)\end{tabular} \\ \hline
xFODE-SR2-PS1 & 148 & 0.0197 $\pm$ (0.0015) & 148 & 0.1271 $\pm$ (0.0032) & 148 & 9.6841 $\pm$ (0.3818) & 148 & 0.1845 $\pm$ (0.0222) & 282 & \begin{tabular}{@{}c@{}}$y_1$: 0.0790 $\pm$ (0.0041)\\$y_2$: 0.0729 $\pm$ (0.0044)\end{tabular} \\ \hline

% PS2 is old PS1
xFODE-SR1-PS2 & 148 & 0.0263 $\pm$ (0.0022) & 148 & 0.2300 $\pm$ (0.0074) & 148 & 10.6379 $\pm$ (0.8213) & 108 & 0.1993 $\pm$ (0.0382) & 282 & \begin{tabular}{@{}c@{}}$y_1$: 0.1083 $\pm$ (0.0125)\\$y_2$: 0.1229 $\pm$ (0.0139)\end{tabular} \\ \hline
xFODE-SR2-PS2 & 148 & 0.0208 $\pm$ (0.0015) & 148 & 0.1249 $\pm$ (0.0045) & 148 & 9.9384 $\pm$ (0.4458) & 108 & 0.1817 $\pm$ (0.0146) & 282 & \begin{tabular}{@{}c@{}}$y_1$: 0.0802 $\pm$ (0.0083)\\$y_2$: 0.0740 $\pm$ (0.0109)\end{tabular} \\ \hline

% PS3 is old PS2
xFODE-SR1-PS3 & 148 & 0.0176 $\pm$ (0.0032) & 148 & 0.2032 $\pm$ (0.0042) & 148 & 10.5588 $\pm$ (0.6652) & 108 & 0.1903 $\pm$ (0.0222) & 282 & \begin{tabular}{@{}c@{}}$y_1$: 0.1192 $\pm$ (0.0171)\\$y_2$: 0.1291 $\pm$ (0.0147)\end{tabular} \\ \hline
xFODE-SR2-PS3 & 148 & 0.0180 $\pm$ (0.0012) & 148 & 0.1293 $\pm$ (0.0043) & 148 & 10.0630 $\pm$ (0.3803) & 108 & 0.1880 $\pm$ (0.0394) & 282 & \begin{tabular}{@{}c@{}}$y_1$: 0.0862 $\pm$ (0.0120)\\$y_2$: 0.0768 $\pm$ (0.0088)\end{tabular} \\ \hline

\end{tabular}}
\begin{tablenotes}
\footnotesize
\scriptsize
\item[*] FODE-SR1 yielded NaN values for 4 seeds. These experiments are excluded from statistical analysis.
\end{tablenotes}
\end{threeparttable}
\end{table*}

\begin{table}[t]
\centering
\caption{Testing performance of NLARX models}
\label{tab:nlarx}
\renewcommand{\arraystretch}{2.0}

% Local reduction of horizontal padding
\begingroup
\setlength{\tabcolsep}{1.25pt}

\begin{adjustbox}{max width=\linewidth}
\begin{tabular}{@{}|l|
                >{\centering\arraybackslash}p{0.16\linewidth}|
                >{\centering\arraybackslash}p{0.16\linewidth}|
                >{\centering\arraybackslash}p{0.16\linewidth}|
                >{\centering\arraybackslash}p{0.16\linewidth}|
                >{\centering\arraybackslash}p{0.20\linewidth}|@{}}
\hline
\multicolumn{1}{|l|}{\textbf{Model$^*$}} & 
\multicolumn{1}{c|}{\textbf{Two-Tank}} & 
\multicolumn{1}{c|}{\textbf{Hair Dryer}} &
\multicolumn{1}{c|}{\textbf{MR Damper}} &
\multicolumn{1}{c|}{\textbf{EV Battery}} & 
\multicolumn{1}{c|}{\textbf{Steam Engine}} \\ \hline

\multirow{2}{*}{SN} &
[3 3 1] & [5 5 0] & [5 5 1] & [2 2 1] & [4 4 1] \\ \cline{2-6}
& 0.0193 & 0.0707 & 5.3760 & 0.1470 & \makecell{$y_1$: 0.0774\\$y_2$: 0.0813} \\ \hline

\multirow{2}{*}{TE} &
[3 3 1] & [10 10 0] & [20 20 0] & [3 3 0] & [4 4 0] \\ \cline{2-6}
& 0.0575 & 0.1318 & 8.1335 & 0.2771 & \makecell{$y_1$: 0.1945\\$y_2$: 0.1760} \\ \hline

\multirow{2}{*}{SVM} &
[5 5 0] & [10 10 0] & [1 1 0] & [20 20 0] & [5 5 1] \\ \cline{2-6}
& 0.1972 & 0.0790 & 16.8883 & 0.7832 & \makecell{$y_1$: 0.1377\\$y_2$: 0.1188} \\ \hline

\multirow{2}{*}{NN} &
[7 7 1] & [5 5 0] & [5 5 0] & [2 2 0] & [5 5 0] \\ \cline{2-6}
& 0.0254 & 0.0875 & 7.5247 & 0.1440 & \makecell{$y_1$: 0.1036\\$y_2$: 0.1231} \\ \hline

\multicolumn{6}{p{\linewidth}}{
\footnotesize
\scriptsize
* For each model, the regressor dimensions $[m_y, m_u, m_d]$ (output lag, input lag, and input delay) and the corresponding RMSE values are provided for each dataset.
} 
\end{tabular}
\end{adjustbox}
\endgroup
\end{table}

\section{Conclusion and Future Work}

This paper introduces the xFODE, which offers interpretable models for SysID. In xFODE, states are constructed from outputs within SR2 to maintain physical meanings. The xFODE framework models state derivative through additive single-input FLSs, revealing input-wise contributions to state dynamics. xFODE provides transparent mapping of FLSs by shaping their antecedent spaces with structured PSs. With these PSs, xFODE ensures that at most two consecutive rules are activated, thereby resulting in clean rule semantics. Results on benchmark datasets reveal that xFODE achieves comparable performance with NODE/FODE using different PSs depending on the dataset. Overall, xFODE effectively enhances the interpretability of input and antecedent spaces while maintaining competitive SysID performance.

Although xFODE improves the interpretability of states and antecedents, the rule consequents remain ambiguous. In future work, we will analyze the rule consequent to improve the interpretability of the learned model further.

\section*{Acknowledgment}
The authors acknowledge using ChatGPT to refine the grammar and enhance the expression of English.

\bibliographystyle{IEEEtran}
\bibliography{IEEEabvr,cites}

 % \addtolength{\textheight}{-20cm}   % This command serves to balance the column lengths
                                  % on the last page of the document manually. It shortens
                                  % the textheight of the last page by a suitable amount.
                                  % This command does not take effect until the next page
                                  % so it should come on the page before the last. Make
                                  % sure that you do not shorten the textheight too much.

%%%%%%%%%%%%%%%%%%%%%%%%%%%%%%%%%%%%%%%%%%%%%%%%%%%%%%%%%%%%%%%%%%%%%%%%%%%%%%%%

%%%%%%%%%%%%%%%%%%%%%%%%%%%%%%%%%%%%%%%%%%%%%%%%%%%%%%%%%%%%%%%%%%%%%%%%%%%%%%%%

%%%%%%%%%%%%%%%%%%%%%%%%%%%%%%%%%%%%%%%%%%%%%%%%%%%%%%%%%%%%%%%%%%%%%%%%%%%%%%%%

\end{document}